# Frequency-domain Blind Quality Assessment of Blurred and Blocking-artefact Images using Gaussian Process Regression model

Maryam Viqar[1], Athar A. Moinuddin[1], Ekram Khan[1] and M. Ghanbari[2]

[1] Department of Electronics Engineering, Z. H. College of Engineering & Technology, AMU, Aligarh
[2] School of Electronics and Computing, University of Essex, Colchester, UK

**ABSTRACT**

Most of the standard image and video codecs are block-based and depending upon the compression ratio the compressed images/videos suffer from different distortions. At low ratios, blurriness is observed and as compression increases blocking artifacts occur. Generally, in order to reduce blockiness, images are low-pass filtered which leads to more blurriness. Also, in bokeh mode images they are commonly seen: blurriness as a result of intentional blurred background while blocking artifact and global blurriness arising due to compression. Therefore, such visual media suffer from both blockiness and blurriness distortions. Along with this, noise is also commonly encountered distortion. Most of the existing works on quality assessment quantify these distortions individually. This paper proposes a methodology to blindly measure overall quality of an image suffering from these distortions, individually as well as jointly. This is achieved by considering the sum of absolute values of low and high-frequency Discrete Frequency Transform (DFT) coefficients defined as sum magnitudes. The number of blocks lying in specific ranges of sum magnitudes including zero-valued AC coefficients and mean of 100 maximum and 100 minimum values of these sum magnitudes are used as feature vectors. These features are then fed to the Machine Learning (ML) based Gaussian Process Regression (GPR) model, which quantifies the image quality. The simulation results show that the proposed method can estimate the quality of images distorted with the blockiness, blurriness, noise and their combinations. It is relatively fast compared to many state-of-art methods, and therefore is suitable for real-time quality monitoring applications.

## 1. INTRODUCTION

Many techniques inspired by biological systems of living beings are employed in tasks of developing artificially intelligent systems, data processing, computation methods, optimization algorithms, etc. As human beings are the end-users in a wide range of applications, hence many-a-times techniques are required that closely replicate the human biological systems. One such application is Image Quality Assessment (IQA) where the methods are designed in a way to perceive the quality of visual media according to the Human Visual System (HVS). Major functionalities of neurons in the visual cortex such as frequency decomposition and divisive normalization transform of visual signals are replicated by such quality assessment methods [1].

### 1.1 MOTIVATION:

Today, images and videos are popular means of information sharing through social media platforms like Instagram, Facebook, YouTube, etc. Furthermore, there is an exponential increase in mobile/Internet data traffic worldwide and it is estimated that by the year 2022, the visual contents will consume about 80% of total Internet traffic [2]. Despite being the main source of information sharing in modern digital era, the most prevalent images and videos are likely to get contaminated with multiple distortions as they undergo through different stages like acquisition, transmission, compression, processing, etc. These operations introduce noises during acquisition and transmissions, blocking artifacts and blurriness during JPEG/MPEG compression, blurriness during filtering or processing, etc. The joint effect

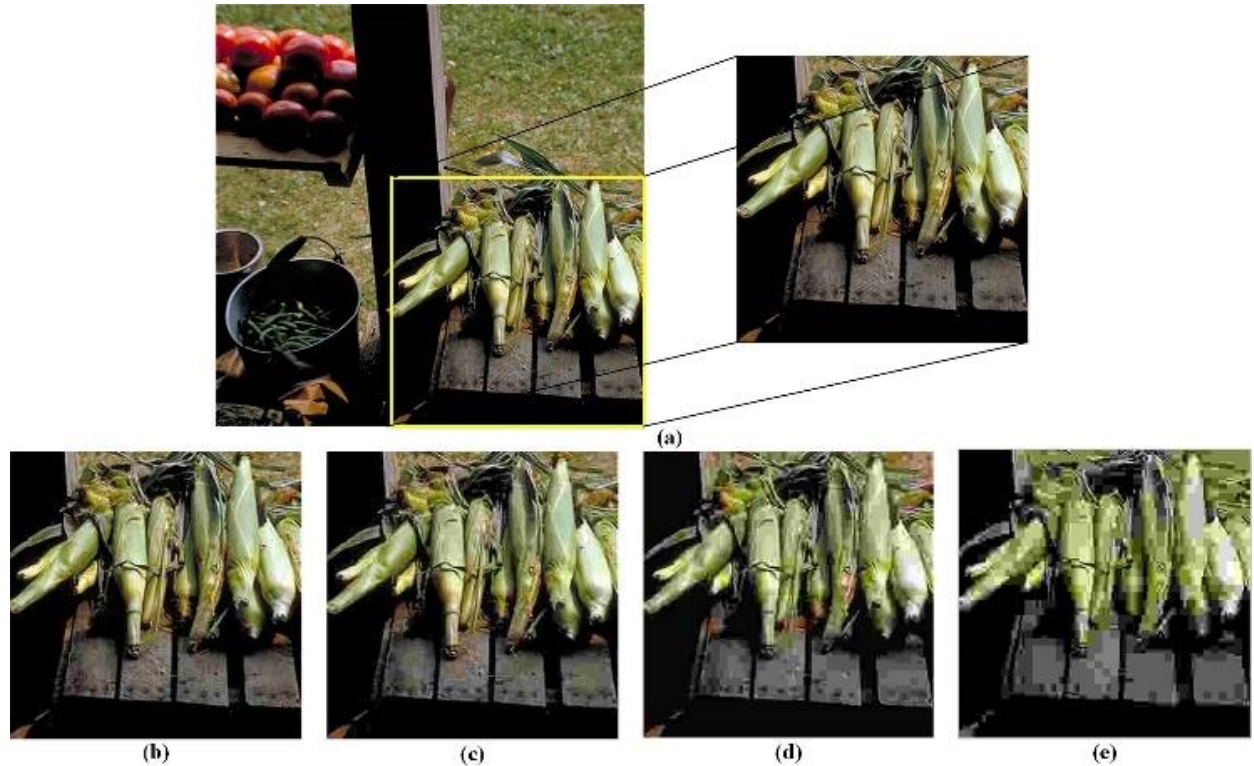

***Fig.1*** *(a) original image CSIQ database [5], JPEG compressed images at various compression ratios (CR):-(b) CR=20, (c) CR=40, (d) CR=65 and (e)CR=120*

of the multiple distortions on image quality is perceptually different from the effect caused by single distortion in images. Therefore, it becomes important to develop IQA method for multiple distortions which have recently gained significant attention.

The image quality can be assessed either subjectively or objectively. Despite being highly consistent, the subjective quality evaluation requires a large number of human observers and therefore it is an expensive and time-consuming process and therefore not suitable for real-time applications. Alternatively, objective image quality assessment techniques do not require any human intervention and are faster. The objective quality evaluation techniques can be classified into three categories: full-reference (FR) which requires the original image to estimate the quality of the target image; reduced reference (RR), which does not require the reference image as a whole, rather uses some of its features; and no-reference (NR) methods, where the quality is being estimated without the reference (or original) image. For most real time applications, the NR (also known as blind) image quality assessment technique is the most suitable approach.

To facilitate the fast exchange of visual information under unprecedented traffic growth, the use of efficient image/video compression techniques becomes unavoidable. Most of the efficient compression techniques are lossy and introduce some degree of distortions as evident from Fig. 1. The highly compressed images/videos are likely to be more distorted, so a tradeoff between compression efficiency and image/video quality needs to be properly managed. It requires a fast and accurate blind quality assessment technique so that the trade-off between quality and degree of compression can be managed on a real-time basis. In coded images/videos, two commonly encountered distortions are the blockiness and blurriness artefacts. The blocking artefacts are observed in block-based JPEG (Joint Photographic Experts Group) and H.26x/MPEG (Moving Picture Experts Group) coded images/videos when a significant difference in quantized DC coefficients of adjacent blocks occurs. On the other hand, coded images appear to be blurred when either the image is coded at low compression ratios causing the removal of the high frequency components, or when the processing like low pass filtering (or loop filtering in standard video coders) are employed to remove the blocking artefacts (specifically in video coding). These two compression related distortions i.e. blockiness and blurriness go hand in hand in images/videos. It has been observed that initially at very low compression rate, first blurriness appears and as the rate is further stepped up, blocking artefact becomes visible [3]. Some authors have exploited such relation between blockiness and blurriness, for designing more efficient image/video quality assessment (IQA) tool. For example, if reference image or its partial information is available, then by measuring the gain and loss at block boundaries, one can design a very efficient IQA [4]. To verify this observation, an image from CSIQ database [5] is compressed at different compression ratios (CR) and corresponding decoded images are shown in Fig. 1. For better visualization a portion of image shown in Yellow Square is cropped and enlarged. The pristine image of Fig. 1(a) is compressed at CR= 20, 40, 65 and 120 which are shown in Figs. 1(b)-(e) respectively. Initially at low compression ratio (CR=20) the image of Fig. 1(b) tends to lose the sharpness, as fine lines of corn skin are disappearing when compared with the original image. As the CR is increased to 65, more blurriness is introduced. When the compression ratio becomes very high (CR=120), blocking artefacts becomes prominent as apparent from Fig. 1(e). From this example, it is visually evident that block-based compression techniques introduce both blurriness as well as blockiness. Thus, any compressed image/video has a mixture of both distortions. Another scenario where a combination of these two distortions appear in coexistence is when loop-filtering is employed, especially in the video encoder loop. Moreover, Additive White Gaussian noise (AWGN) is also commonly experienced distortion which generally affects the perceptual quality of images by introducing a high-frequency random component. The motivation of the proposed work is to develop an IQA method that can accurately estimate the quality of images distorted with blockiness, blurriness and/or noise (individually or jointly).

**1.2 RELATED WORK:**

The existing blind or NR-IQA (no-reference IQA) methods can be categorized into two broad groups, namely: distortion specific methods and generic methods. Sometimes, the nature of distortion may be known e.g., quantization, distortions due to camera motion, inherent noise sources inside camera, low-pass filtering, etc. However, in many cases the source and nature of distortions are anonymous. The distortion specific methods, as the name suggests, are designed to measure image quality under a specific type of single distortion. They can be employed when the image is exposed to a single known type of distortion. In contrast, generic methods are designed to assess different kinds of individual distortions.

In the distortion specific IQA methods, we consider here the methods specifically designed for blurriness, blocking artefacts or noise. The quality of blurred images can be assessed by measuring the spread of edges [6]- [8], by measuring the just noticeable blur (JNB) [9]- [11], or by computing kurtosis on the Discrete Cosine Transform (DCT) blocks [12]. In [13], authors use Maximum Local Variation (MLV) for each pixel to estimate the sharpness of an image. The MLV's are assigned weights and their standard deviation is used to measure the degree of sharpness, which is then used to quantify the blurriness. The concept of Structural Similarity Index Measure (SSIM) was exploited by Zhang et al. in [14] to detect blurriness in images while re-blurring it. It is a two-step framework in which first the image is filtered by a Gaussian low pass filter (GLPF) and then the corresponding changes in histogram are analyzed to determine extent of blurriness. In [15], authors have used the Exact Zernike Moments & Gradient Magnitude (EZMGM) for establishing a blurriness metric. H. Liu et al. have proposed a blur metric in [16] which utilizes edge blur and neural network and works well on images having intentionally blurred background.

Similarly, numerous techniques have been developed to measure the distortion due to blocking artefacts [17]-[24]. Most of these techniques exploit the frequency-domain characteristics, mainly the Discrete Fourier Transform (DFT) [17], [21], and Discrete Cosine Transform (DCT) [18], [20]. The combination of DFT and DCT has also been exploited to assess the quality of JPEG compressed images [22], where authors use the number of zero-valued coefficients in DCT transformed blocks, weighted by DFT-generated quality maps as features to assess the degree of blockiness. Though the concept is simple and effective, but it has poor accuracy for most of the standard databases. A gradient-based technique is proposed in [23], in which for each 8×8 pixels non-overlapping block, two parameters namely luminance change (across adjacent blocks) and degree of blockiness (evaluated by adding the horizontal and vertical derivatives) are determined to measure the image quality. In general, this method has a better accuracy, but it fails to accurately estimate the quality if blocks are surrounded with flat regions. Zhu et al. [24] proposed a technique based on quality factor (QF) and recompression of already compressed JPEG images. The methods proposed in [25]- [28] were developed for quality assessment of noisy images.

Most of the methods mentioned above though have good performance on either of distortions individually but they become very inefficient for images distorted with more than one type of distortions. As in the present era, the images are likely to be exposed to commingled distortions, diverse kind of generic IQA methods are required which can estimate image quality in presence of multiple known/unknown distortions with high accuracy. Some of the relevant generic methods are reviewed next.

A full reference method capable of assessing image quality distorted by both blurriness and blockiness using frequency domain approach is proposed in [3]. A number of generic blind image quality assessment methods exists like [29] - [39]. The majority of these methods are based on Natural Scene Statistics (NSS) of the image [29] - [34], [37], [39], etc. which rely on the fact that when an image is exposed to distortions, its statistical parameters get changed, which can be utilized to estimate the degree of distortion in the image. The NSS features may be extracted either in spatial domain or in transform domain or in a combination of both. Moorthy and Bovik proposed a technique named DIIVINE (Distortion Identification-based Image Verity and INtegrity Evaluation) [31] which assess the image quality in two steps. Identification of distortion in the first step is followed by the quantification of degree of distortion in the second step. It uses wavelet transform to obtain statistical features and Gaussian Scale Mixture for modelling the features. This method uses a large feature size (a total of 88 features), which obviously makes it too complex. BLIINDS2 (BLind Image Integrity Notator using DCT Statistics) [32] is a method that uses fewer number of DCT-based NSS features. Here the DCT coefficients of each block are modelled using Generalized Gaussian density (GGD) model and the parameters of GGD model are fed to Bayesian regression model to quantify the image quality. Though BLINDS2 [32] method has better efficiency as compared to DIIVINE [31] method, the large processing time due to the use of non-linear sorting of block-based features in BLIINDS2 [32] makes it slow and imposes restriction for its use in many applications [29]. The BRISQUE (Blind / Referenceless Image Spatial Quality Evaluator) [29] method is a spatial domain NSS based method, which uses Mean Subtracted Contrast Normalized (MSCN) coefficients of image to develop a quality assessment metric. Though BRISQUE is one of the fastest (in terms of computational time) IQA method, but it performs poorly for specific type of images as in TID2013 database [40]. Xue et al. [33] have utilized Gradient magnitude (GM) and Laplacian of Gaussian (LoG) jointly to study the effect of local luminance changes and intensity variations respectively on the image quality, as they are closely related to the HVS. This method measures the quality of images exposed to blocking artefacts accurately but has relatively lower accuracy for blurred images. A combination of both spatial and transform domain features have been utilized in FRIQUEE (Feature Maps–Based Referenceless Image Quality Evaluation Engine) method [37]. A large number of features across different color spaces makes it complex and time-consuming. Li et al. in [39] proposed a blind method based on log-contract distribution to exploit spatial and directional correlations in images. This method suffers from data dependency and its accuracy for cross-database validations is not satisfactory. The methods discussed here are ML based, whereas learning free generic methods like NIQE [41] and IL-NIQE [42] also exit. The work of M H Pinson [43] draws attention towards assessment of visual media from the perspective of consumer applications. The methods like BRISQUE [29], NIQE [41], IL-NIQE [42], etc. have been tested for multiple distortions, but found to have poor performance for these distortions as well as for unseen data. Additionally, few multiple distortion methods [44]-[47] have also been proposed to assess the image quality in presence of two or more commingled type of distortions. GWH-GLBP [44] was proposed by Li et al. to assess multiple distortions in images. It is based on structural information wherein local binary pattern (LBP) from the gradient map of the test image is used to form the weighted histograms. Miao et. al [45] have proposed another method based on LBP in Phase Congruency domain to take into account low level features. It also employs gradient magnitude as a weighing factor. Six-step blind metric (SISBLIM) [46] is a training free metric based on joint effect's prediction, and HVS based fusion model. A significant effort has been made by VQEG (Video Quality Experts Group) [48] for quality assessment of visual media under projects like

MOAVI (Monitoring of Audio-Visual quality by key Indicators) [49], [50] and NORM (No Reference Metric). A set of key indicators to define the service quality of audio-visual signals were proposed in the MOAVI project. The NORM project is concerned with research and development and sharing of resources of NR metrics.

Most of generic IQA methods developed to measure image quality in presence of multiple distortions either suffer with large time-complexity or with poor accuracy. In context of multiple distorted images, many state-of-art methods give poor performance when tested on images exposed to multiple distortions, due to the joint effect of various distortions. Furthermore, multiple distortions IQA area could not receive much attention possibly due to the lack of benchmark databases as well as its challenging nature. Hence, there is a need of an IQA method which can assess quality of images having multiple distortions with higher accuracy and lower complexity.

To overcome the drawbacks of IQA methods that measure image quality under multiple distortion conditions, we propose a DFT-based technique to estimate the quality of images distorted with blocking artefact, blurriness, noise individually or in combination of two or more of these distortions. Though DFT and MSCN coefficients individually have been widely utilized in many IQA methods, less efforts have been made to combine the benefits of both these methods. We believe that the novelty of our proposed method lies in developing an IQA method by utilizing the DFT of MSCN coefficients. Most of the existing MSCN methods use the univariate probabilistic models for fitting distributions which are insufficient in apprehending correlation properties. Another limitation arises when only neighbouring pixels are considered, as distortion generally corrupts multiple pairs simultaneously and hence are inappropriate for complex types of distortions. To overcome these limitations, a unique method based on MSCN and DFT is proposed. To be specific, the novelty of this work lies in: (i) utilization of the combination of the MSCN with DFT which helps in obtaining a highly decorrelated low-dimensional space (ii) considering the effect of distortion on multiple pixels (rather than only neighbouring pixels) from a suitable sized block of coefficients. The proposed technique first computes the MSCN coefficients of input image. The original image as well as MSCN coefficients are then transformed into frequency domain using 8×8 block-based DFT. The DFT coefficients of each block are then divided into low, medium and high frequency bands. The high and low frequency band coefficients are summed into a sum parameter, obtained from DFT of original image and DFT of MSCN coefficients to extract features. For normalized sum parameters, number of blocks lying in the specific ranges are counted including the zero-valued coefficients along with the mean of 100 largest and 100 smallest values of these sum parameters are computed to form the feature vectors. All these features are then fed to Gaussian Process Regression (GPR) model [51] with exponential kernel for quality estimation. The proposed method has much better accuracy over the methods developed to measure these distortions separately.

The organization of paper is as follows: Section 2 briefly reviews the DFT and MSCN, which are being used in the proposed method. Section 3 gives the details of the proposed methodology. Section 4 includes the simulation results and discussions on these results. Finally, the paper is concluded in Section 5.

## 2. BACKGROUND

Since proposed IQA method exploits DFT- domain features of image combined with DFT of MSCN coefficients of input image, hence, in this section DFT and MSCN coefficient computations and their characteristics are briefly reviewed.

*2.1 Discrete Fourier Transform (DFT):*

It is a well observed fact that distortions in images affect their frequency distribution characteristics. Therefore, the relative changes in strengths of high and low frequency components of images can be used to determine if an image is distorted with blocking artefact, noise and blurriness. These variations can easily be captured by DFT, making it an obvious choice in this work. Fig. 2 shows a set of eight images taken from LIVE [52] database where Figs. 2(a) and 2(i) are the original images while Figs. 2(b)-(d) are JPEG compressed placed in order of decreasing quality (increasing DMOS) values. Similarly, Figs. 2(j)-(l) are images exposed to Gaussian blurriness, arranged in order of increasing DMOS. Block-based 8×8 DFT is applied to each block, and magnitude of DFT coefficients of a block at correspondingly same location (shown with yellow square) in each image are also shown in Figs. 2(e)-(h) and Figs. 2(m)-(p) corresponding to images shown in Figs. 2(a)-(d) and Figs. 2(j)-(l) respectively. From these figures it can be observed that distribution of DFT coefficients vary with the nature and degree of degradation (artefact) present in images. These fluctuations in magnitude of DFT coefficients can be utilized for quality assessment of images. The 2-D DFT $F(u,v)$ can be computed as in Eqn. (1), where I(m,n) is the gray-scale image (or image block), having N×M pixels.

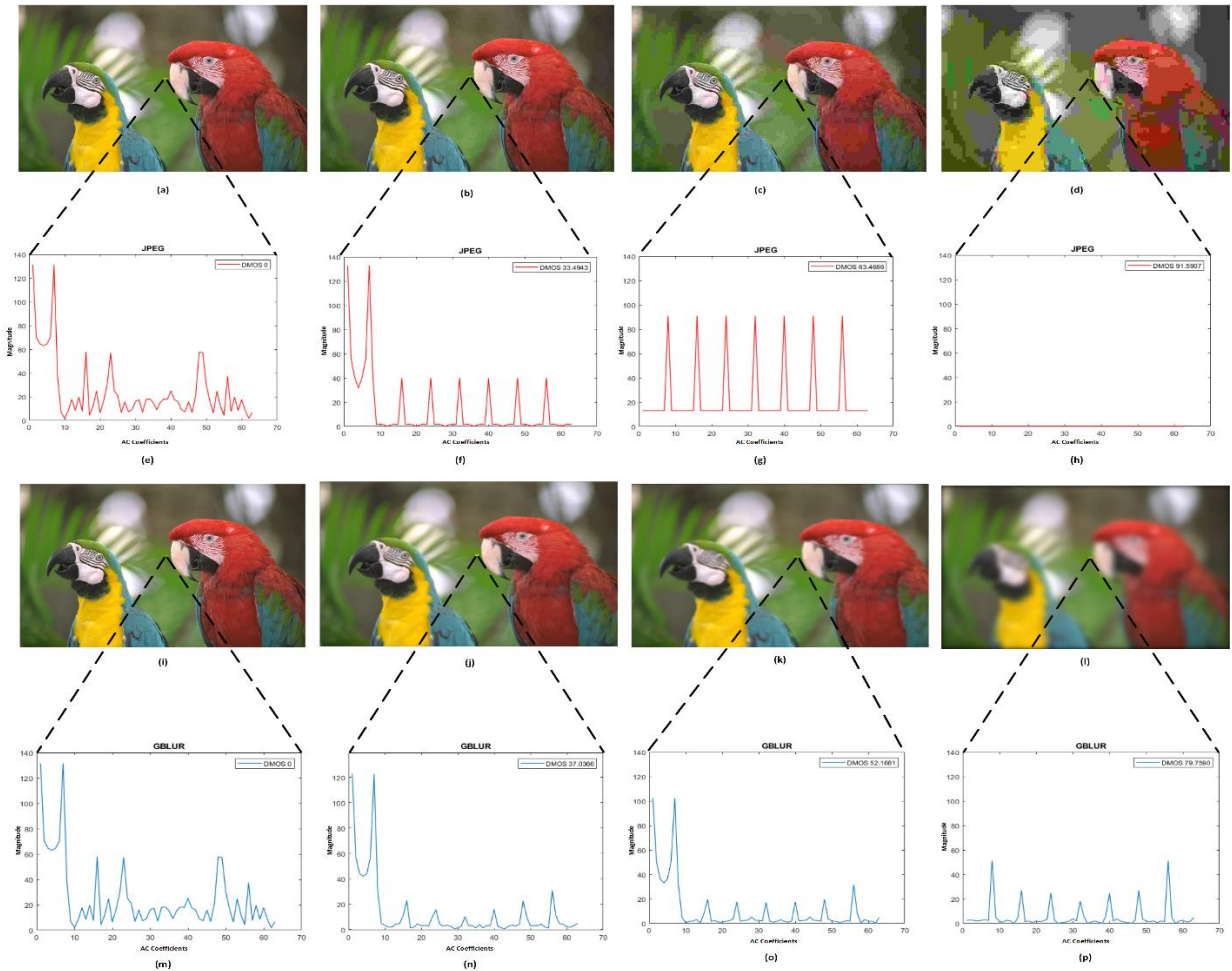

***Fig.2*** *(a)-(d) JPEG and (i)-(l) GBLUR (Gaussian Blur) type of images from LIVE [52] database with corresponding plots for DFT Coefficients(e)-(h) and (m)-(p) respectively.*

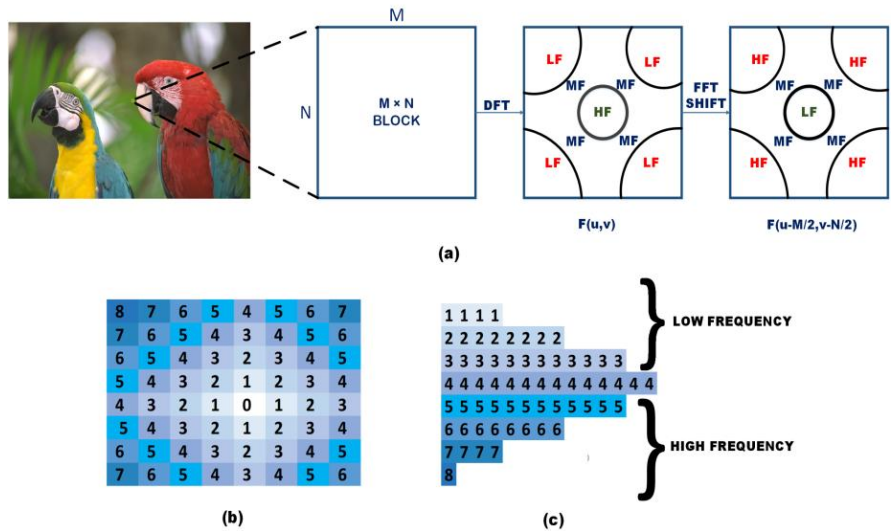

***Fig.3*** *(a) Block level representing of DFT coefficients followed by frequency shifting (b) block level representation of DFT components indexed in terms of increasing frequency using Manhattan distance (c) Segregation of high and low frequency sub-bands by separating into indexed rows using Manhattan distance*

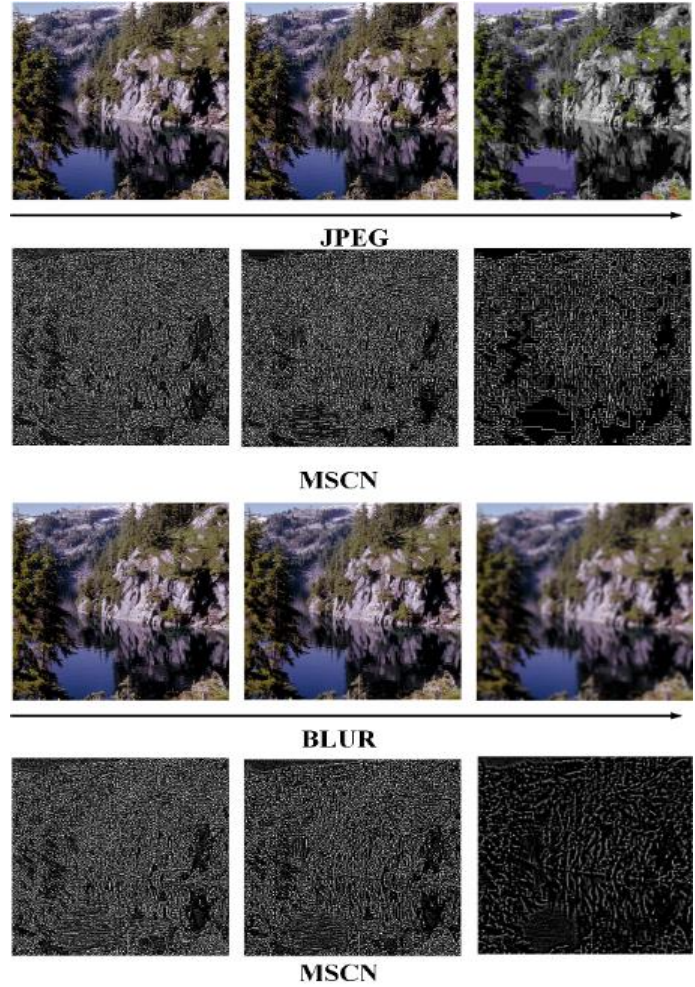

***Fig.4*** *(a) Row one contains JPEG compressed images and Row 3 blurred images from CSIQ Database [5]. Corresponding MSCN images of row1 are shown in row 2 and that for blur images of row 3 are shown in row 4.*

$$F(u,v) = \sum_{m=0}^{M-1}\sum_{n=0}^{N-1} I(m,n) e^{-j2\pi\left(\frac{u}{M}m+\frac{v}{N}n\right)} \tag{1}$$

In order to visualize spectrum symmetry of DFT, generally the spectrum is circularly shifted to bring the zero frequency (u=0, v=0) at the center, which is also treated as origin of spectral plane, as depicted in Fig. 3(a). The frequency shifting can be represented mathematically as in Eqn. (2):

$$F_1(u,v) = F\left(u-\frac{M}{2}, v-\frac{N}{2}\right) \tag{2}$$

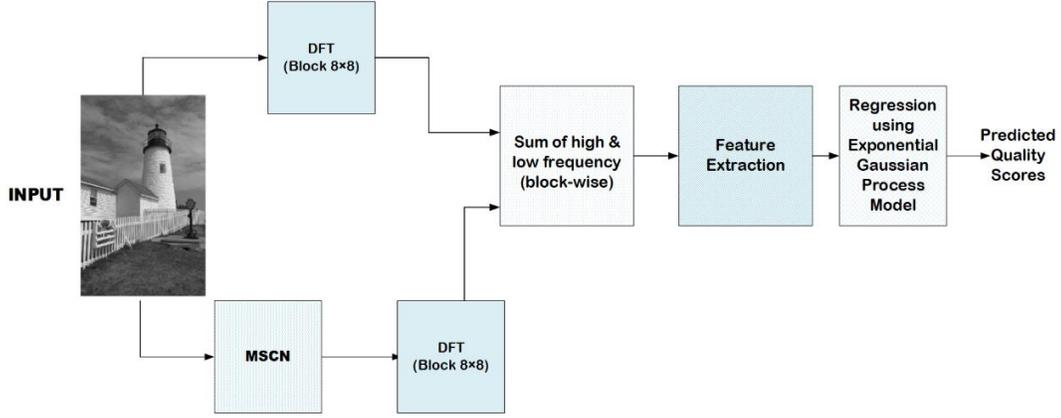

*Fig.5 Block diagram of the proposed IQA method*

where $F_1(u, v)$ is the frequency shifted (shifted by M/2 horizontally and N/2 vertically) version of $F(u, v)$ defined in Eqn. (1). After frequency shifting, coefficients of each block are indexed using Manhattan Distance between DC coefficient and other coefficients of the block. Accordingly, the index '$i$' of frequency coefficient located at (u, v) is computed as:

$$i = |u_0 - u| + |v_0 - v| \tag{3}$$

where $(u_0, v_0)$ represents co-ordinates of the DC coefficient at center (0,0). According to Eqn. (3), the frequency coefficients are numerically indexed in terms of increasing frequency using Manhattan distance, irrespective of direction they occur. The DFT coefficients of an 8×8 block can then be divided into low, medium and high frequencies depending upon their frequency index '$i$', as shown in Fig.3 (b). Here DFT coefficients of an 8×8 block is indexed from 0-8 depending upon their Manhattan distance. The DC component lying at the center is indexed as '0'. As one moves away from the center, the frequency index '$i$' increases. In the proposed scheme, we will use this indexing to segregate low and high frequency DFT coefficients.

*2.2 Mean Subtracted Contrast Normalized Coefficients (MSCN):*

Human Visual System (HVS) is habituated to its natural surroundings. This feature is embodied in Natural Scene Statistics (NSS) Model [53]. Over past decades, researchers have employed NSS for establishing several IQA metrics [29] - [34], [37], [39], [41], [42], etc. showing the relevance of NSS to human perceptions. An important parameter derived from NSS model is MSCN [29]. Mathematically, it can be represented as:

$$\hat{I}(m, n) = \frac{I(m, n) - \mu(m, n)}{\sigma(m, n) + 1} \tag{4}$$

In Eqn.(4) gray-scale image is denoted as $I(m, n)$, m $\in$ {1,2,…..M} (M denotes image height), n $\in$ {1,2,…..N} (N denotes image width) are spatial indices.

$$\mu(m, n) = \sum_{k=-K}^{K} \sum_{l=-L}^{L} w_{k,l} I_{k,l}(m, n) \tag{5}$$

$$\sigma(m, n) = \sqrt{\sum_{k=-K}^{K} \sum_{l=-L}^{L} w_{k,l} (I_{k,l}(m, n) - \mu(m, n))^2} \tag{6}$$

Eqn.5 and Eqn.6 calculate the local mean and contrast respectively. Here, w={$w_{k,l}|k = -K, ….., K, l = -L, … L$} is a Gaussian weighting function. It is sampled out to 3 standard deviations where K=L=3 and rescaled to unity volume. In terms of symmetry, it has 2D circular symmetry. This kind of normalization performs almost similar operation as performed in the visual brain wherein each neuronal response is normalized by the energy of its adjacent neuronal responses [1]. Cells of visual cortex give non-linear response especially in cases of high contrast where linearity-based models fail [54]. Subtraction of mean from the image helps in eliminating the effect of uneven illumination. The variance field itself is capable of showing local edges and high contrast details. The combined normalization process brings the intensity values in a normalized range; leaving behind a highly decorrelated homogenous image with low energy edges having important structural information [29]. This normalization helps in removing of spatial redundancies, giving a homogenous appearance. It works on similar pattern like the non-linear response of visual neurons. Fig.4 shows two sets of images chosen from CSIQ Database [5] where Row 1 shows a set of JPEG compressed images while Row 3 consists of images distorted by blurriness arranged in increasing order of distortion. Rows 2 and 4 show the corresponding images after MSCN operation for JPEG and blurred images of Rows 1 and 3 respectively. The presence of residual boundaries having low energy can be visualized in second and fourth rows.

## 3. PROPOSED IQA METHODOLOGY

The block diagram of the proposed IQA algorithm to estimate the quality of images separately/jointly corrupted with blockiness, blurriness and noise is shown in Fig. 5. It works as follows. First a gray-level input image is normalized to get corresponding MSCN image. The pristine image as well as MSCN coefficients are then transformed into frequency domain using 8×8 DFT. The frequency coefficients of every block in each image are divided into low and high frequency bands. After frequency domain transformation, four sum parameters are obtained by adding the high and low frequency coefficients for DFT blocks obtained from gray and MSCN images. For these sum parameters, we determine the Normalization Factor (NF) using the variance of block count for sum-parameters. After normalizing the sum-parameters with NF, the number of blocks having values of these parameters in specific ranges are counted. These block-counts are used as features. Generally, frequency domain distribution of undistorted natural images is peaky with heavy tails while for noisy images it is otherwise. As noise affects the high frequency components, the mean of hundred largest and hundred smallest values of high frequency sum parameters are also computed as features. For mapping the features on quality scale, Gaussian Process Regression model [51] with exponential kernel is used. Various components of the proposed method are described below.

*3.1 Motivation for combining DFT with MSCN:*

The motivation to use DFT along with MSCN in the proposed work can be justified in terms of the number of zero-valued coefficients in a DFT block. The zero-valued DFT coefficients was first proposed by Golestaneh et al. [22] for estimating the degradation in a block of pixels in an image, which is also used in this work to access the quality of images. For this purpose, we have considered three (an original and two distorted) versions of the same image as shown in Figs. 6(a), 6(d) and 6(g) respectively. These images are taken from LIVE database [52], and their corresponding DMOS values are 0, 47.28 and 83.55 respectively. That is image of Fig. 6(g) has more distortions (poor quality) than the image of Fig. 6(d). Each image is then 8×8 DFT transformed (either directly or after MSCN normalization). Then zero-valued coefficients are mapped with black pixel and non-zero valued coefficients with white pixels. The directly DFT transformed and mapped images are shown in Figs. 6(b), 6(e) and 6(h) corresponding to the images of Figs. 6(a), 6(d) and 6(g) respectively. Similarly, Figs. 6(c), 6(f) and 6(i) show the black & white mapped DFT coefficients of MSCN normalized versions of images of Figs. 6(a), 6(d) and 6(g) respectively. It can be observed from these figures that as distortion increases, the number of zero-valued DFT coefficients (black pixels) increases. Furthermore, it can be observed by comparing Figs. 6(e) with 6(f) and Figs. 6(h) with 6(i), that DFT with MSCN can differentiate the local distortions much better than DFT only coefficients.

Further to justify why DFT coefficients with and without MSCN are considered in the proposed work, we have considered six sets of JPEG images (original and distorted versions) from the LIVE database [52] where original six images (Image 1, Image2, …, Image 6) are shown in Fig. 7. Each set of images (original as well as distorted) are transformed using 8×8 DFT and the number of zero-valued coefficients in transformed image are computed. In order to get the idea of correlation between zero-valued DFT coefficients and corresponding image quality (in terms of DMOS value), the number of zero-valued DFT coefficients for each set of images are plotted against DMOS value (taken from the database) of the corresponding image, as shown in Fig. 7(a). It can be observed from this figure that for first 4 images (Image 1-4), the number of zero-valued coefficients are highly correlated with DMOS, but there exists a poor correlation between these parameters for Image 5 and Image 6. It may be noted that these two images (Image 5 and Image 6) have locally blurred regions. Thus, the number of zero-valued ac coefficients in DFT transformed images fail to correlate with image quality of locally distorted images. However, when the same set of images are DFT transformed after MSCN normalization, then the number of zero-valued coefficients have good correlation with DMOS of corresponding images for locally distorted image, as shown with red-color in Fig. 7 (b), but it has poor correlations for globally distorted images (Image 1-4). It should be noted for these images (Image 1-4), number of zero-valued DFT coefficients (for gray image) has good correlation as evident from Fig. 7 (a). Hence, in order to have better quality estimate for every distorted image, features from DFT transformed images, both with and without MSCN should be computed. These features are then passed to a Machine Learning algorithm for further processing.

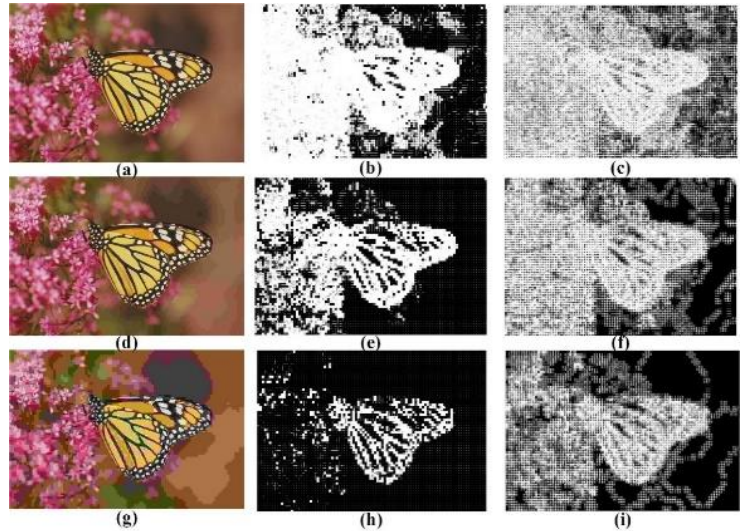

***Fig.6*** *Three JPEG images from LIVE [52] database with DMOS (a) 0, (d) 47.28 and (g) 83.55, DFT on gray scale images is shown by (b), (e) and (h) respectively, DFT applied on MSCN coefficients is shown by (c), (f) and (i).*

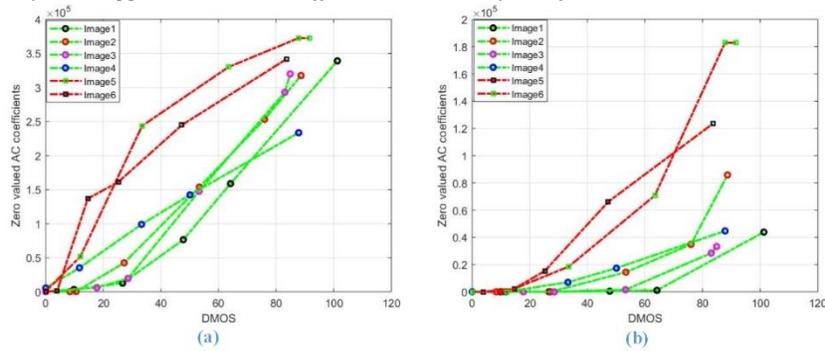

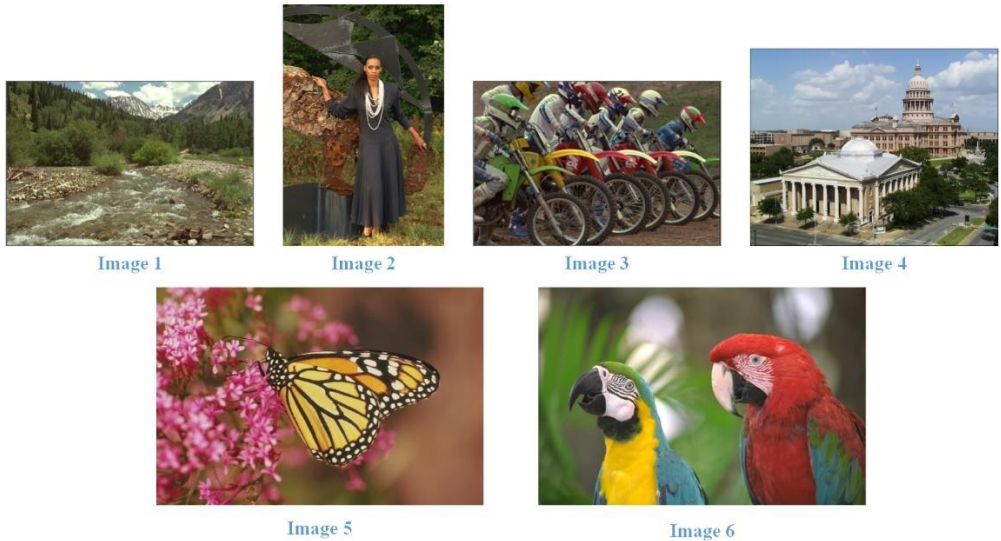

***Fig.7*** *Image 1,2,3,4,5 and 6: pristine images from LIVE [52] database. (a) Plot for Zero-valued coefficients for DFT on gray scale images (b) Plot for Zero-valued coefficients for DFT on MSCN image (set of 1,2,3,4,5 and 6 containing pristine and all distorted versions from JPEG LIVE [52] Database).*

*3.2 Frequency segregation:*

As discussed earlier, DFT is applied to the input image on a block-by-block basis (size 8×8 pixels). Parallel to this, MSCN coefficients are extracted from the image to take into account the characteristics of NSS model inspired by HVS. The computed MSCN coefficients are also transformed to frequency domain by employing the block-based DFT. In this way, two (with and without MSCN) frequency domain representations of the image are obtained. Let the DFT transform of gray-scale input image ($I(m,n)$) and its MSCN normalized version ($\hat{I}(m,n)$) be $F_g(u,v)$ and $F_m(u,v)$ respectively. It is observed that the energy distribution in low and high frequency components of image is likely to be different for different types of distortions in image. Furthermore, since the energy contained in a DFT coefficient is equal to the square of its absolute value, therefore we consider only the magnitude of DFT coefficients. Let the absolute values of $F_g(u,v)$ and $F_m(u,v)$ be denoted as $|F_g(u,v)|$ and $|F_m(u,v)|$ respectively.

Further, as discussed previously in Section 2.1 that in order to segregate the low and high frequencies, the DFT coefficients of a block are indexed on the basis of Manhattan distance from the DC coefficient of the block, as defined in Eqn. (3). The coefficients with the same frequency index are grouped together as shown in Fig. 3(c), and all AC coefficients having frequency indices 1-3 are labelled as low-frequency (LF) coefficients, index 4 as middle frequency coefficient (MF) and indices 5-8 as high-frequency (HF) coefficients. In the proposed work, only coefficients belonging to LF and HF bands will be used for feature extraction. If $F_{k,g} = \{F_g(u,v)\}_k$ and $F_{k,m} = \{F_m(u,v)\}_k$ represent the set of all DFT coefficients of a DFT block (without and with MSCN respectively) having $k^{th}$ frequency index, then LF and HF band of image in DFT domain (without and with MSCN) are the set of all DFT coefficients with frequency indices 1-3 and 5-8, and can be represented according to Eqns. (7)-(10) for each block:

$$LF_g = \bigcup_{k=1}^{3}\{F_{k,g}\} \tag{7}$$

$$HF_g = \bigcup_{k=5}^{8}\{F_{k,g}\} \tag{8}$$

$$LF_m = \bigcup_{k=1}^{3}\{F_{k,m}\} \tag{9}$$

$$HF_m = \bigcup_{k=5}^{8}\{F_{k,m}\} \tag{10}$$

*3.3 Four Sum-parameters:*

In order to determine the feature vectors, we propose to sum the magnitudes of all DFT coefficients belonging to sets $LF_g, HF_g, LF_m,$ and $HF_m$, as defined in Eqns. (11)-(14) respectively, to be considered as feature to quantify the quality of distorted images.

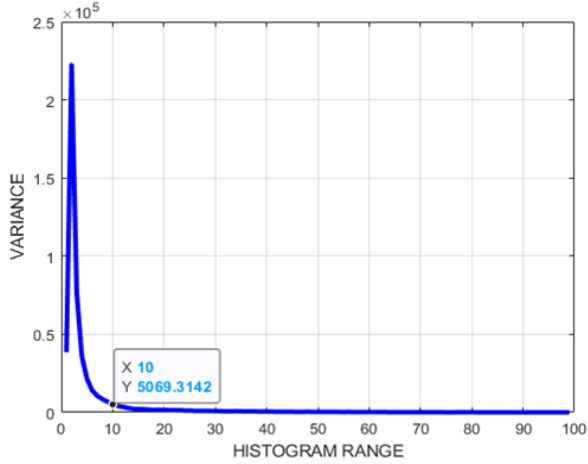
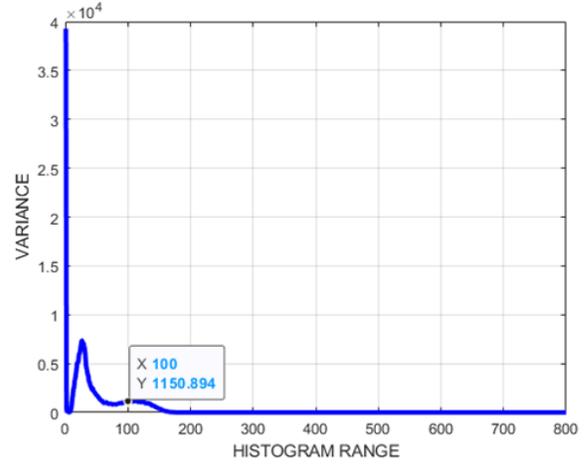
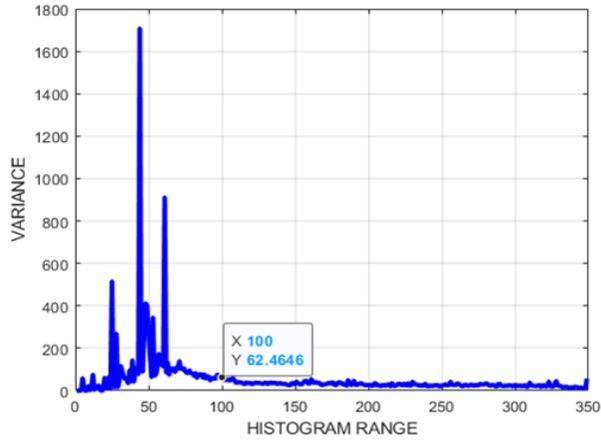
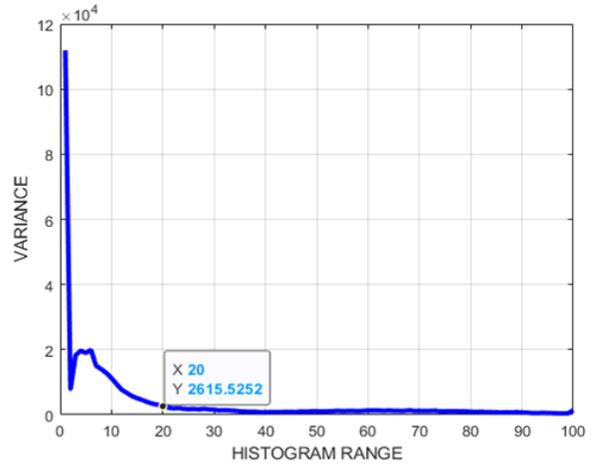

**Fig.8** Variance distribution for sum parameters (a) $S_g^{LF}$ (b) $S_m^{LF}$ (c) $S_g^{HF}$ (d) $S_m^{HF}$

$$S_g^{LF} = \cup_{all\ blocks} \sum_{k=1}^{3}|F_{k,g}| \tag{11}$$

$$S_g^{HF} = \cup_{all\ blocks} \sum_{k=5}^{8}|F_{k,g}| \tag{12}$$

$$S_m^{LF} = \cup_{all\ blocks} \sum_{k=1}^{3}|F_{k,m}| \tag{13}$$

$$S_m^{HF} = \cup_{all\ blocks} \sum_{k=5}^{8}|F_{k,m}| \tag{14}$$

Since the range in which the values in set of sum parameters $S_g^{LF}$, $S_g^{HF}$, $S_m^{LF}$, and $S_m^{HF}$ lie is different and they also depend on the nature of distortions. For further processing, each of sum-parameters should be normalized on the scale of 0 to 1, which is discussed next.

*3.4 Normalization Factor:*

The set of four parameters related to low-frequency and high-frequency components of DFT (with and without MSCN) defined in Eqns. (11) - (14) above, may have any values depending upon image content and nature/degree of distortions. In order to get an idea about the range of these four sum parameters, a set of 100 images from GBLUR (Gaussian Blur) and JPEG compressed subsets of LIVE Database [52] are selected randomly. For each of the images, the set of four sum parameters defined in Eqns. (11) - (14) were evaluated. The number of blocks that exist in different ranges of these sum-parameters are computed. The number of times a specific bin value (which corresponds to number of blocks) occurs for all 100 images are counted individually for each sum parameter and variance of these counts is shown in Fig. 8. The variance of values (corresponding to each bin) is utilized to determine the maximum range of each of the sum-parameters. The bin size for Fig.8 (a) is 100 whereas for remaining parameters (Figs. 8(b), 8(c), and 8(d)), the bin-size is 1. From Fig. 8, it is clear that most of images have values of individual sum parameters from the sets $S_g^{LF}$, $S_g^{HF}$, $S_m^{LF}$, and $S_m^{HF}$ lying in the range of 0-10 (bin-size 100, Fig. 8(a)), 0-100 (Fig. 8(c)), 0-100 (Fig. 8(b)) and 0-20 (Fig. 8(d)) respectively. In order to bring them in common range of 0 to 1, there is a need of normalization. The maximum range for each sum parameter is considered as normalization factor (NF), so that each sum parameter can be expressed on the scale of 0 to 1. The normalization factors (NF) for the four sets of sum-parameters $S_g^{LF}$, $S_g^{HF}$, $S_m^{LF}$, and $S_m^{HF}$ are 1000, 100, 100 and 20 respectively. After determining the sum-parameters, they are divided by their respective NF to ensure that normalized value of these parameters remain on the scale of 0 to 1. It may be noted that the normalization factor ensures that both types of distortions are defined within a common scale.

*3.5 Feature extraction:*

Finally, features are extracted as follows. Since all sum parameters are normalized on the scale of 0 to 1, where zero values of each of sum-parameter corresponds to zero-valued AC coefficients in the image. Therefore, the number of blocks with zero-value of each of sum parameter are used as four features (one for each sum parameter). The entire range (except zero value) is divided into four equal parts 0-0.25 (excluding 0), 0.25-0.50, 0.50-0.75 and 0.75-1.0. The number of blocks having each of sum-parameters in each of four ranges will constitute 4×4=16 features. As images may be of different sizes, therefore percentage of blocks rather than absolute number of blocks for each of sum-parameter lying in different ranges are considered as feature vectors. Since there are a total of four sum-parameters and each is divided into 5 classes, therefore we get 20 features in this manner. These 20 feature vectors are defined in Table 1, and these are the numbers of blocks belonging to each class divided by total number of blocks in an image. The features $f_1$, $f_2$, $f_3$, $f_4$ and $f_5$ are percentages of blocks having normalized values of sum-parameter from set $S_g^{LF}$ as 0, in the range of 0-0.25, 0.25-0.50, 0.50-0.75 and 0.75-1.0 respectively. Similarly, $f_6$-$f_{10}$ correspond to sum-parameter set $S_m^{LF}$, $f_{11}$-$f_{15}$ to $S_g^{HF}$ and $f_{16}$-$f_{20}$ to $S_m^{LF}$, $f_{11}$-$f_{15}$ to $S_g^{HF}$ and $f_{16}$-$f_{20}$ to $S_m^{HF}$. In addition to this, four more features (f21-f24) are extracted by calculating the mean of hundred largest and hundred smallest values from the set of sum parameters $S_g^{HF}$ and $S_m^{HF}$ respectively. Hence, a total of 24 features are derived from the four sum-parameter sets.

*3.6 Regression stage:*

ML-based Regression module is used for mapping the feature vectors to their respective quality scores. Gaussian process regression (GPR) is a kernel-based supervised approach and non-parametric in nature. There are many of its variants depending on kernel function used, such as linear, exponential, square exponential, a combination of multiple kernels, etc. In the present work, exponential GPR module is used to map frequency domain NSS features of the image to its quality score.

Gaussian Process (GP) is a stochastic process that can be represented as a distribution over function $\xi$ such that it maps input feature space $\mathbf{F}$ (in our case feature vectors) to output space $\mathbf{D}$ (here space consisting of DMOS of images). It can be modelled using the mean $m(f)$ and the covariance $C(f, f')$ (also works as kernel of GP) as defined in Eqn.15 and Eqn.16 respectively.

$$m(f) = E[\xi(f)] \tag{15}$$

$$C(f, f') = E[(\xi(f) - m(f))(\xi(f') - m(f'))] \tag{16}$$

In Eqn.16 $f$ and $f'$ are elements of set $\mathbf{F}$ and E is the expectation operator. In the proposed framework, the input space $\mathbf{F}$ is the set of features $f1$ to $f20$ listed in Table I. GP is then a non-linear function of $m(f)$ and $C(f, f')$ and is defined in Eqn. 17 as follows:

$$\xi(f) \sim GP(m(f), C(f, f')) \tag{17}$$

It achieves the mapping of inputs $f_i \in \mathbb{F}$ to an output space $d_i \in \mathbb{D}$ by imposing restriction that a group of input-output training points $\{(f_i, d_i) | i = 1, \ldots, n\}$ are known, in priory in order to obtain the posterior distribution. The output $d$ of the GP model is a noisy observation represented as:

$$d = \xi(f) + \acute{\varepsilon} \tag{18}$$

where $\acute{\varepsilon}$ is the additive Gaussian noise $N(0, \sigma^2)$. To obtain the predicted output DMOS values from testing dataset, this posterior distribution is used. The covariance, being the kernel of GP model is also known, whose hyper parameters are optimized during the training phase. The joint distribution of training output $d_{tr}$ and test output $d_p$ is to be defined as:

$$\begin{bmatrix} d_{tr} \\ d_p \end{bmatrix} \sim N \left( 0, \begin{bmatrix} C(F,F) + \sigma_n^2 I & C(F, F_*) \\ C(F, F) & C(F, F_*) \end{bmatrix} \right) \tag{19}$$

Where F and $F_*$ are feature vectors for training and test data respectively. The standard deviation of noise is given by $\sigma_n^2 I$ where I is the identity matrix (N-dimensional). To obtain the predictive function, the observation $d_p$ is conditioned on $d_{tr}$ such that

$$d_p | F, d_{tr}, F_* \sim N(\overline{d_p}, V(d_p)) \tag{20}$$

where $\overline{d_p} = C(F_*, F)[C(F, F) + \sigma_n^2 I]^{-1} d_{tr}$, $\tag{21}$

$$V(d_p) = C(F_*, F_*) - C(F_*,) [C(F, F) + \sigma_n^2 I]^{-1} C(F, F_*) \tag{22}$$

GP predicts the output $d_p$ using this predictive distribution. The test output which is the predicted DMOS is $d_p$ as defined in Eqn. 20. For more details on GP one can refer to [51]. Regression Learner App from the Machine Learning toolbox of MATLAB 18a is used in this work for the training and validation of GP model.

## 4. RESULTS AND DISCUSSIONS

In this section, we perform an analysis of the proposed model to examine its performance in accordance with human perception on the relevant well-known single-distortion based databases LIVE [52], CSIQ [5], TID2013 [40] and KADID-10k [55]. To test the accuracy of state-of-art and proposed methods on multiple distortions, LIVEMD [56], IVL MD [57], [58], MDID2013 [46] and MDID [59] datasets and for Bokeh mode images IBBI database [16], [60] are used. The methods considered for comparisons, include distortion specific such as Ferzli [9], EZMGM [15], Golestaneh [22], Zhan [23], Zhu [24], H. Liu [16] as well as general purpose methods such as DIIVINE [31], BLIINDS2 [32], CORNIA [38], BRISQUE [29], GMLoG [33], NFERM [34], FRIQUEE [37], BJLC [39], NIQE [41] and IL-NIQE [42]. Some methods designed for multiply distorted images namely GWH-GLBP [44], Miao [45], Zhou [47] and SISBLIM [46] specifically, are also included for fair comparison. These methods can be grouped into two classes: ML-based and non-ML based. The IQA methods DIIVINE [31], BLIINDS2 [32], CORNIA [38], BRISQUE [29], GMLoG [33], NFERM [34], FRIQUEE [37], BJLC [39], H. Liu [16], GWH-GLBP [44], Miao [45] and Zhou [47] are ML-based methods, whereas NIQE [41], IL-NIQE [42], Golestaneh [22], Zhan [23], Zhu [24], Ferzli [9], Bahrami [13], EZMGM [15] and SISBLIM [46] are non-ML-based methods.

*4.1 Databases and Evaluation criteria:*

Several databases involving different types of images exposed to varying types and levels of distortions are available. The details of some of the used databases are as follows:

- LIVE [52] consists of 29 reference images, 233 are afflicted to JPEG compression, 174 are exposed to GBLUR and 145 images have undergone AWGN type of distortion referred here as LIVE.
- The CSIQ [5] database has 30 reference images, along with 150 distorted images in JPEG, BLUR and AWGN sub-datasets respectively referred here as CSIQ.

- TID2013 [40] database has a total of 24 types of distortions with 25 reference images out of which 120 are JPEG-compressed, 120 are exposed to GBLUR and 120 afflicted with AWGN referred as TID2013 in this work. Five images in each sub-set are not included as they are not natural images.
- KADID-10k [55] is the largest single distortion-based dataset consisting of 10,125 distorted images obtained from 81 reference images. This work includes GBLUR, JPEG compression and AWGN subsets consisting 405 distorted images respectively.
- LIVEMD database [56] has a total of 405 distorted images out of which 270 are multiply distorted. It consists of 15 reference images which are exposed to three distortion types (Gaussian blur, JPEG compression, and white Gaussian noise). In addition to individual subsets exposed to single type of distortion, the database has two sub-sets exposed to multiple distortions of type: (i) Gaussian blur followed by JPEG compression and (ii) Gaussian Blur followed by White Noise (WN). These two subsets are used to assess the quality of multiply distorted images and is referred here as LIVEMD.
- IVL MD database [57], [58] has 10 original images and it is divided into two subsets of multiply distorted images consisting of 750 multiply distorted images. The database consists of two subsets: 1) Blur-JPEG where each original image undergoes 7 levels of Gaussian blur, then each resultant blur image undergoes 5 levels of JPEG compression and 2) Noise-JPEG where each reference image is afflicted with 10 levels of Gaussian Noise and each noisy image is exposed to 4 levels of JPEG compression.
- The Multiply Distorted Image Database 2013 (MDID2013) database [46] comprises of 324 multiply distorted images, generated by successively distorting 12 original images by three types of distortions (GBLUR, JPEG, and WN).
- The Multiply Distorted Image Database (MDID) [59] (different from MDID2013) contains 20 reference and 1600 multiply distorted images by introducing five types of distortions: Gaussian noise, Gaussian blur, contrast change, JPEG, and JPEG2000 compression. In this work, Gaussian noise, Gaussian blur and JPEG afflicted images have been used for analysis and referred to as MDID.
- IBBI [16], [60] database has 12 reference images and 60 distorted images having intentionally blurred background.

For performance evaluation different criteria have been used mainly: Spearman rank order correlation coefficient (SROCC), Pearson linear correlation coefficient (PLCC), Kendall's Rank-order Correlation Coefficient (KROCC) and Root Mean Squared Error (RMSE). The range for SROCC, KROCC and PLCC is [-1, 1], whereas RMSE is a non-negative real number. These parameters depict the correlation between predicted and subjective scores. A lower RMSE and a higher SROCC/PLCC/KROCC depict a good correlation with the subjective judgments. Due to non-linear human response system, there may be non-linearity in subjective and predictive scores. Hence, a logistic function is used for logistic mapping of scores before computing PLCC and RMSE [61]. The function can be defined as in Eqn. 23 where x and f(x) are the predicted scores before and after the regression, $\beta_1, \beta_2, \beta_3, \beta_4,$ and $\beta_5$ are regression model parameters.

**Table 2** SROCC, PLCC, KROCC and RMSE values for JPEG, GBLUR and WN type of distortions on LIVE, CSIQ and TID2013 databases. Best two results are marked using bold face.

| METHOD | LIVE | | | | CSIQ | | | | TID2013 | | | |
|---|---|---|---|---|---|---|---|---|---|---|---|---|
| | SROCC | PLCC | KROCC | RMSE | SROCC | PLCC | KROCC | RMSE | SROCC | PLCC | KROCC | RMSE |
| DIIVINE [31] | 0.831 | 0.826 | 0.685 | 15.24 | 0.832 | 0.839 | 0.642 | 0.146 | 0.675 | 0.635 | 0.507 | 1.016 |
| BLIINDS2 [32] | 0.907 | 0.914 | 0.737 | 11.08 | 0.849 | 0.873 | 0.653 | 0.131 | 0.751 | 0.781 | 0.564 | 0.820 |
| NIQE [41] | 0.924 | 0.916 | 0.755 | 10.98 | 0.856 | 0.870 | 0.663 | 0.132 | 0.736 | 0.741 | 0.529 | 0.898 |
| IL-NIQE [42] | 0.902 | 0.908 | 0.722 | 11.62 | 0.871 | 0.874 | 0.682 | 0.155 | 0.863 | 0.868 | 0.616 | 0.655 |
| CORNIA [38] | 0.945 | 0.951 | 0.812 | 8.75 | 0.916 | 0.881 | 0.763 | 0.153 | 0.899 | **0.938** | 0.686 | **0.563** |
| BRISQUE [29] | 0.957 | 0.960 | 0.819 | 7.77 | 0.914 | **0.936** | 0.752 | **0.095** | 0.832 | 0.852 | 0.663 | 0.688 |
| GMLoG [33] | 0.962 | 0.957 | 0.823 | 7.79 | 0.926 | 0.914 | 0.788 | 0.113 | **0.931** | 0.926 | **0.763** | 0.603 |
| NFERM [34] | 0.942 | 0.955 | 0.810 | 8.48 | 0.921 | 0.933 | 0.776 | 0.101 | 0.908 | 0.911 | 0.691 | 0.615 |
| FRIQUEE [37] | 0.963 | 0.953 | 0.823 | 8.55 | **0.939** | 0.895 | **0.795** | 0.149 | 0.926 | 0.845 | 0.755 | 0.629 |
| BJLC [39] | **0.968** | **0.963** | **0.834** | **7.61** | 0.923 | 0.928 | 0.785 | 0.102 | **0.941** | 0.885 | **0.783** | 0.618 |
| PROPOSED | **0.979** | **0.981** | **0.861** | **5.67** | **0.932** | **0.948** | **0.794** | **0.096** | 0.922 | **0.928** | 0.748 | **0.599** |

**Table 3** SROCC for JPEG, GBLUR and WN type of individual distortions on LIVE, CSIQ and TID2013 databases. Best two results are marked using bold face.

| METHOD | JPEG | | | BLUR | | | WN | | |
|---|---|---|---|---|---|---|---|---|---|
| | LIVE | CSIQ | TID2013 | LIVE | CSIQ | TID2013 | LIVE | CSIQ | TID2013 |
| *Golestaneh [22]* | 0.954 | 0.923 | 0.891 | - | - | - | - | - | - |
| *Zhan [23]* | 0.961 | 0.940 | 0.936 | - | - | - | - | - | - |
| *Zhu [24]* | 0.961 | **0.952** | **0.952** | - | - | - | - | - | - |
| *Ferzli [9]* | - | - | - | 0.943 | 0.886 | 0.857 | - | - | - |
| *EZMGM [15]* | - | - | - | 0.958 | 0.932 | 0.944 | - | - | - |
| *Bahrami [13]* | - | - | - | 0.954 | 0.925 | 0.936 | - | - | - |
| DIIVINE [31] | 0.914 | 0.888 | 0.826 | 0.938 | 0.856 | 0.943 | 0.979 | 0.940 | 0.906 |
| BLIINDS2 [32] | 0.951 | 0.898 | 0.775 | 0.926 | 0.924 | 0.858 | 0.941 | 0.922 | 0.741 |
| NIQE [41] | 0.943 | 0.882 | 0.880 | 0.939 | 0.906 | 0.835 | 0.961 | 0.835 | 0.852 |
| IL-NIQE [42] | 0.944 | 0.904 | 0.883 | 0.924 | 0.867 | 0.864 | 0.977 | 0.866 | 0.904 |
| CORNIA [38] | 0.945 | 0.906 | 0.891 | **0.960** | 0.914 | 0.922 | 0.979 | 0.941 | 0.935 |
| BRISQUE [29] | 0.963 | 0.904 | 0.869 | 0.947 | 0.919 | 0.893 | 0.981 | 0.957 | 0.900 |
| GMLoG [33] | 0.965 | 0.916 | 0.923 | 0.938 | 0.915 | 0.929 | 0.978 | 0.943 | 0.946 |
| NFERM [34] | 0.967 | 0.922 | 0.911 | 0.948 | 0.897 | 0.926 | 0.980 | 0.938 | 0.931 |
| FRIQUEE [37] | 0.955 | 0.934 | 0.912 | 0.956 | **0.942** | 0.945 | 0.975 | 0.941 | **0.950** |
| BJLC [39] | **0.968** | **0.950** | 0.946 | 0.954 | 0.930 | **0.968** | **0.986** | **0.962** | **0.958** |
| PROPOSED | **0.979** | 0.948 | **0.958** | **0.972** | **0.944** | 0.951 | **0.984** | **0.960** | 0.947 |

**Table 4** PLCC for JPEG, GBLUR and WN type of individual distortions on LIVE, CSIQ and TID2013 databases. Best two results are marked using bold face.

| METHOD | JPEG | | | BLUR | | | WN | | |
|---|---|---|---|---|---|---|---|---|---|
| | LIVE | CSIQ | TID2013 | LIVE | CSIQ | TID2013 | LIVE | CSIQ | TID2013 |
| *Golestaneh [22]* | 0.970 | 0.954 | 0.938 | - | - | - | - | - | - |
| *Zhan [23]* | 0.978 | **0.963** | 0.962 | - | - | - | - | - | - |
| *Zhu [24]* | 0.971 | **0.984** | 0.962 | - | - | - | - | - | - |
| *Ferzli [9]* | - | - | - | 0.911 | - | 0.852 | - | - | - |
| *EZMGM [15]* | - | - | - | 0.956 | **0.928** | **0.942** | - | - | - |
| *Bahrami [13]* | - | - | - | **0.959** | 0.907 | 0.875 | - | - | - |
| DIIVINE [31] | 0.934 | 0.897 | 0.899 | 0.937 | 0.898 | 0.844 | 0.971 | 0.899 | 0.882 |
| BLIINDS2 [32] | 0.943 | 0.912 | 0.889 | 0.899 | 0.901 | 0.825 | 0.932 | 0.897 | 0.714 |
| NIQE [41] | 0.935 | 0.893 | 0.882 | 0.936 | 0.897 | 0.819 | 0.946 | 0.855 | 0.851 |
| IL-NIQE [42] | 0.942 | 0.908 | 0.876 | 0.918 | 0.885 | 0.868 | 0.974 | 0.858 | 0.917 |
| CORNIA [38] | 0.965 | 0.877 | **0.963** | 0.955 | 0.882 | 0.941 | 0.965 | 0.942 | 0.937 |
| BRISQUE [29] | 0.973 | 0.946 | 0.951 | 0.951 | **0.928** | 0.863 | 0.974 | 0.938 | 0.810 |
| GMLoG [33] | 0.953 | 0.907 | 0.925 | 0.942 | 0.920 | 0.931 | 0.966 | 0.944 | 0.925 |
| NFERM [34] | **0.981** | 0.944 | 0.907 | 0.937 | 0.897 | 0.928 | **0.979** | 0.936 | 0.939 |
| FRIQUEE [37] | 0.947 | 0.885 | 0.813 | 0.949 | 0.905 | 0.881 | **0.982** | 0.947 | **0.955** |
| BJLC [39] | 0.961 | 0.942 | 0.873 | 0.952 | 0.926 | 0.853 | - | - | - |
| PROPOSED | **0.986** | 0.949 | **0.966** | **0.971** | 0.941 | 0.953 | 0.977 | **0.965** | 0.946 |

$$f(x) = \beta_1 \left( \frac{1}{2} - \frac{1}{\exp(\beta_2(x-\beta_3))} \right) + \beta_4 x + \beta_5 \quad (23)$$

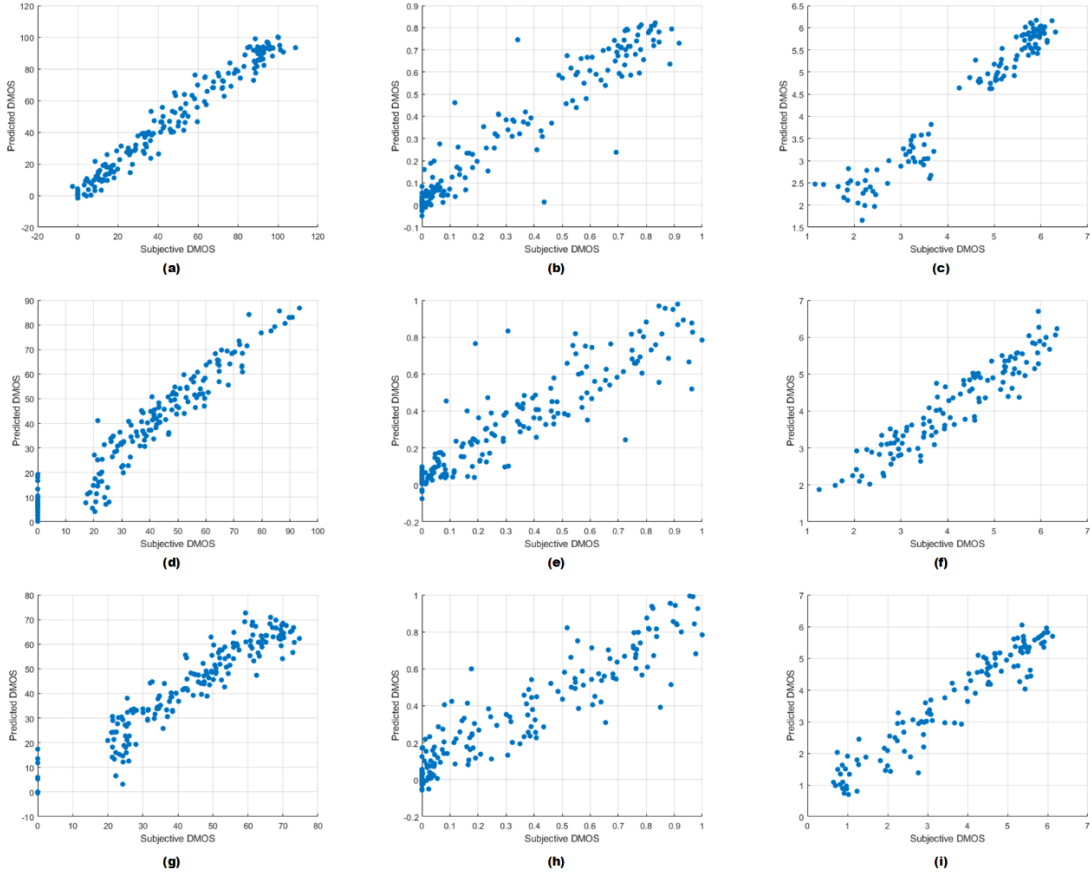

*Fig.9* Plots between standard and predicted scores for different Datasets. (a) LIVE [52] DATASET for JPEG compressed images (b) CSIQ [5] DATASET for JPEG compressed images (c) TID2013 [40] DATASET for JPEG compressed images (d) LIVE [52] DATASET for BLUR type of distortion (e) CSIQ [5] DATASET for BLUR type of distortion (f) TID2013 [40] DATASET for BLUR type of distortion

**Table 5** SROCC and PLCC on KADID-10k for JPEG, GBLUR and WN image datasets and Bokeh mode images on IBBI dataset. Best two performing methods are highlighted using bold face.

| | KADID-10k | | | | | | IBBI | |
|---|---|---|---|---|---|---|---|---|
| | SROCC | | | PLCC | | | SROCC | PLCC |
| | JPEG | GBLUR | WN | JPEG | GBLUR | WN | BLUR | BLUR |
| DIIVINE [31] | 0.766 | 0.787 | 0.625 | 0.805 | 0.794 | 0.648 | 0.978 | 0.955 |
| BLIINDS2 [32] | 0.789 | 0.759 | 0.718 | 0.812 | 0.806 | 0.682 | 0.971 | 0.974 |
| ILNIQE [42] | 0.871 | 0.865 | 0.737 | 0.908 | 0.843 | 0.694 | 0.964 | 0.972 |
| NIQE [41] | 0.848 | 0.872 | 0.829 | 0.928 | 0.893 | 0.826 | 0.978 | 0.965 |
| BRISQUE [29] | 0.789 | 0.814 | 0.583 | 0.810 | 0.828 | 0.588 | **0.980** | **0.984** |
| GMLoG [33] | 0.811 | 0.809 | 0.847 | 0.856 | 0.845 | 0.816 | 0.972 | **0.981** |
| NFERM [34] | 0.865 | 0.884 | 0.864 | 0.916 | **0.938** | 0.857 | 0.984 | 0.972 |
| FRIQUEE [37] | **0.896** | **0.910** | **0.871** | **0.948** | 0.936 | **0.884** | 0.960 | 0.962 |
| H. Liu [16] | - | - | - | - | - | - | - | 0.973 |
| PROPOSED | **0.917** | **0.934** | **0.904** | **0.966** | **0.957** | **0.902** | **0.985** | 0.977 |

*4.2 Performance on Individual Database:*

Performance of the proposed method was analyzed for single distortion on five databases namely LIVE[52], CSIQ[5],

**Table 6** Cross Validation for SROCC comparison of JPEG and BLUR type of distortion. Best performance is marked using bold face.

|  | LIVE:CSIQ | | LIVE:TID13 | |
| --- | --- | --- | --- | --- |
|  | JPEG | GBLUR | JPEG | GBLUR |
| *Golestaneh [22]* | 0.882 | -- | 0.871 | -- |
| *Zhan [23]* | 0.937 | -- | 0.927 | -- |
| *Zhu [24]* | **0.941** | -- | **0.951** | -- |
| *Ferzli [9]* | -- | 0.874 | -- | 0.851 |
| *EZMGM [15]* | -- | 0.922 | -- | 0.925 |
| *Bahrami [13]* | -- | **0.924** | -- | **0.938** |

**Table 7** Cross Validation for SROCC and PLCC comparison of JPEG and BLUR type of distortion. Best two performances are marked using bold face

|  | LIVE:CSIQ (SROCC) | | | LIVE:CSIQ (PLCC) | | | LIVE:TID2013 (SROCC) | | | LIVE:TID2013 (PLCC) | | |
| --- | --- | --- | --- | --- | --- | --- | --- | --- | --- | --- | --- | --- |
|  | JPEG | GBLUR | WN | JPEG | GBLUR | WN | JPEG | GBLUR | WN | JPEG | GBLUR | WN |
| DIIVINE [31] | 0.845 | 0.888 | 0.876 | 0.831 | 0.862 | 0.851 | 0.845 | 0.888 | 0.868 | 0.837 | 0.835 | 0.857 |
| BLIINDS2 [32] | 0.857 | 0.783 | 0.893 | 0.858 | 0.803 | 0.898 | 0.857 | 0.783 | 0.736 | 0.876 | 0.774 | 0.712 |
| CORNIA [38] | 0.895 | **0.913** | 0.750 | 0.887 | 0.854 | 0.763 | 0.895 | **0.913** | 0.736 | 0.882 | 0.905 | 0.738 |
| BRISQUE [29] | 0.889 | 0.872 | 0.899 | 0.905 | 0.871 | 0.882 | 0.889 | 0.872 | 0.820 | 0.871 | 0.856 | 0.844 |
| ILNIQE [42] | 0.899 | 0.857 | 0.848 | 0.891 | 0.874 | 0.836 | 0.867 | 0.834 | 0.885 | 0.868 | 0.840 | 0.863 |
| NIQE [41] | 0.882 | 0.895 | 0.809 | 0.863 | 0.881 | 0.828 | 0.862 | 0.815 | 0.816 | 0.884 | 0.811 | 0.823 |
| GMLoG [33] | 0.895 | 0.911 | 0.896 | 0897 | 0.894 | 0.905 | 0.895 | 0.911 | 0.899 | **0.892** | **0.918** | 0.903 |
| NFERM [34] | 0.913 | 0.883 | 0.916 | **0.918** | 0.882 | **0.917** | 0.913 | 0.883 | **0.905** | 0.853 | 0.891 | **0.922** |
| FRIQUEE [37] | 0.835 | 0.877 | 0.849 | 0.817 | **0.896** | 0.844 | 0.835 | 0.877 | 0.810 | 0.838 | 0.862 | 0.846 |
| BJLC [39] | **0.941** | 0.899 | 0.932 | - | - | - | 0.916 | 0.891 | 0.903 | - | - | - |
| PROPOSED | **0.937** | **0.946** | **0.956** | **0.945** | **0.932** | **0.961** | **0.948** | **0.939** | **0.938** | **0.946** | **0.937** | **0.948** |

KADID-10k [55], IBBI [16], [60] and TID2013 [40] and compared with other state-of-the-art (SoA) methods. The source codes of the considered methods namely DIIVINE [31], BLIINDS2 [32], CORNIA [38], BRISQUE [29], GMLoG [33], NFERM [34], FRIQUEE [37], GWH-GLBP [44], NIQE [41], IL-NIQE [42], Miao [45] and SISBLIM [46] were obtained from respective author's websites and amongst these ML-based methods were trained on the considered datasets in the same way as the proposed method was trained. For each of these methods, including the proposed method, each dataset was divided for training and testing subsets randomly in the ratio of 80%:20% and 50%:50% (for section 4.6 only). The reported results are the median value on the 1000 independent iterations performed for train-test splitting for fair analysis. For training-free methods like NIQE [41] and IL-NIQE [42], 1000 trials were performed on the test set for consistency and the corresponding median value are considered. For performance evaluation of other methods namely BJLC [39], Zhou [47], Golestaneh [22], Zhan [23], Zhu [24], Ferzli [9], EZMGM [15], Bahrami [13], and H. Liu [16] their results were taken either from corresponding original papers or most relevant related articles. Since, Golestaneh [22], Zhan [23], Zhu [24] methods were developed for JPEG distortion, whereas Ferzli [9], EZMGM [15], Bahrami [13] were designed for GBLUR type of distortions, hence the scores for these methods will be reported for corresponding specific distortion type only. For method proposed by Zhou [47], scores are reported only on MDID2013 [46] and LIVEMD [56] as the original paper mentions results only on these two databases.

The experimental results for combined datasets LIVE [52], CSIQ [5] and TID2013 [40] containing all images exposed to GBLUR, JPEG and WN type of distortions are shown in Table 2. The results of the best two performing methods highlighted using bold face shows that the proposed method gives consistently good results compared with other state-of-art methods. Tables 3 and 4 represent prediction monotonicity and prediction accuracy respectively for individually distorted datasets containing images exposed to GBLUR, JPEG and WN type of distortions corresponding to each of the three databases namely LIVE [52], CSIQ [5] and TID2013 [40]. From these results it can be clearly observed that for all databases, the proposed method is always among the best two performing methods except for JPEG compressed images from CSIQ [5] database and WN images from TID2013 databases. The proposed method gives consistently good performance compared to other methods and is comparable with FRIQUEE [37] and BJLC [39] methods. Specifically, for TID2013 [40] much celebrated methods like BRISQUE [29], GMLoG [33] and NFERM [34] have significantly lower performance when PLCC values are compared whereas the proposed method has the best performance for both SROCC and PLCC values.

Furthermore, to test the wider validity of the proposed method, the SROCC and PLCC values are compared with other methods on KADID-10k [55] and IBBI [16], [60] databases and results are shown in Table 5. It may be noted that KADID-10k database used for analysis in this work consists of images exposed to individual type of distortions namely GBLUR, JPEG compression and White Noise, whereas IBBI [16], [60] database consists of intentionally blurred background images. It can be observed from these results that our method maintains its position among best two methods for both the criteria (SROCC and PLCC) and for all the three distortions in KADID-10k database. However, for images from IBBI [16], [60] database, the proposed method still outperforms majority of SoA methods but has slightly inferior performance compared to BRISQUE [29] and GMLoG [33] methods. When compared to H. Liu [16] method which is a quality assessment method primarily developed for IBBI [16], [60] dataset, the proposed method has superior performance. It may be noted that results of [16] are quoted from the original paper itself.

The six scatter plots for predicted and subjective DMOS scores on two distortions and three considered databases shown in Fig. 9 (a)-(f) further demonstrate the prediction accuracy of the proposed method. It may be noted that under the ideal conditions, the best functioning models should have highly linear and tightly grouped scattering patterns. From the scatter plots of Fig. 9, the linear performance can be seen for both types of distortions in three databases, except for CSIQ [5] database with GBLUR type of distortion. The cause may be the effect of distortion on color component as we are considering only the gray-scale version of the images. Furthermore, highly clustered points can be visualized for JPEG and GBLUR type of images in LIVE [52] and TID2013 [40] databases shown in Fig.9 in accordance with SROCC (0.979, 0.972), (0.958, 0.951) and PLCC (0.986, 0.971), (0.966, 0.953) as listed in Table 3 and Table 4 respectively. The superior performance of proposed method for all performance metrics makes it a preferable algorithm for accurate quality assessment of images independently/jointly distorted with JPEG and GBLUR artefacts.

*4.3 Performance on Cross-Database:*

In the previous section, only database specific and distortion specific results were presented. In order to evaluate the performance of proposed method independent of the specific database along with better generalization capability, cross-database validation is performed by training and testing the methods on different databases with similar distortions. This ensures that learnings from one database are not specific to it and can be used on different databases too.

In this work, the images of the LIVE [52] database are used for training while CSIQ [5] and TID2013 [40] were used for testing. This generates two cross-database combinations (train-test) denoted as LIVE:CSIQ and LIVE:TID2013 respectively and corresponding results (SROCC and PLCC) for each cross-dataset and for each distortion are shown in Table 7. For distortion specific methods only SROCC have been listed in Table 6. It can be observed that the proposed method has best performance for both the distortions across the two databases. BJLC [39] performs slightly better than the proposed method for JPEG type of distortion on LIVE:CSIQ cross-validation for SROCC values, whereas SROCC for JPEG compressed images from LIVE:TID2013 of the proposed method shows better results. Zhu [24] also outperforms on JPEG type of distortion when SROCC values are compared for LIVE:TID2013, but the same is not true for LIVE:CSIQ based cross validations. For WN type of distortion also, the proposed method mostly outperforms other methods, showing good correlation scores. For individual database performance, BJLC [39] was giving consistently good performance but for cross-database validation it has significantly lower correlation scores, making it a highly database and training dependent model. Clearly, the proposed model stands apart as a highly robust and database independent model.

*4.4 Performance analysis on multi-distortion databases:*

**Table 8** SROCC and KROCC comparison of multiple distortion images for the combination of JPEG, GBLUR and WN. Best two performances are marked using bold face.

|  | LIVEMD | | IVL MD | | MDID2013 | | MDID | |
|---|---|---|---|---|---|---|---|---|
|  | SROCC | KROCC | SROCC | KROCC | SROCC | KROCC | SROCC | KROCC |
| GWH-GLBP [44] | 0.944 | 0.805 | 0.889 | 0.717 | 0.908 | 0.719 | **0.892** | **0.708** |
| Miao [45] | **0.955** | **0.810** | **0.941** | **0.798** | **0.921** | **0.744** | **0.845** | **0.661** |
| NIQE [41] | 0.858 | 0.726 | 0.837 | 0.644 | 0.614 | 0.456 | 0.657 | 0.467 |
| IL-NIQE [42] | 0.900 | 0.763 | 0.872 | 0.691 | 0.707 | 0.538 | 0.693 | 0.493 |
| BLIINDS2 [32] | 0.887 | 0.753 | 0.803 | 0.617 | 0.808 | 0.623 | 0.773 | 0.568 |
| BRISQUE [29] | 0.912 | 0.782 | 0.815 | 0.621 | 0.819 | 0.642 | 0.766 | 0.559 |
| GMLoG [33] | 0.833 | 0.704 | 0.901 | 0.731 | 0.838 | 0.657 | 0.771 | 0.568 |
| NFERM [34] | 0.898 | 0.759 | 0.871 | 0.688 | 0.851 | 0.674 | 0.803 | 0.603 |
| SISBLM [46] | 0.907 | 0.771 | 0.848 | 0.672 | 0.894 | 0.686 | 0.744 | 0.538 |
| ZHOU [47] | 0.943 | - | - | - | 0.907 | - | - | - |
| PROPOSED | **0.957** | **0.813** | **0.950** | **0.816** | **0.912** | **0.727** | 0.842 | 0.648 |

**Table 9** PLCC and RMSE comparison of multiple distortion images for the combination of JPEG, GBLUR and WN. Best two performances are marked using bold face

|  | LIVEMD | | IVL MD | | MDID2013 | | MDID | |
|---|---|---|---|---|---|---|---|---|
|  | PLCC | RMSE | PLCC | RMSE | PLCC | RMSE | PLCC | RMSE |
| GWH-GLBP [44] | 0.948 | 5.873 | 0.916 | 6.736 | 0.895 | 0.025 | **0.897** | **1.048** |
| Miao [45] | **0.959** | **5.484** | 0.941 | 6.378 | 0.910 | 0.024 | 0.828 | 1.618 |
| NIQE [41] | 0.887 | 8.933 | 0.837 | 7.256 | 0.648 | 0.037 | 0.678 | 2.766 |
| IL-NIQE [42] | 0.914 | 7.536 | 0.871 | 7.018 | 0.709 | 0.034 | 0.733 | 2.217 |
| BLIINDS2 [32] | 0.904 | 7.864 | 0.803 | 7.884 | 0.842 | 0.028 | 0.785 | 2.015 |
| BRISQUE [29] | 0.928 | 7.141 | 0.815 | 7.697 | 0.833 | 0.029 | 0.781 | 2.059 |
| GMLoG [33] | 0.872 | 9.164 | 0.901 | 6.841 | 0.830 | 0.028 | 0.791 | 1.956 |
| NFERM [34] | 0.917 | 7.459 | 0.872 | 6.943 | 0.875 | 0.027 | 0.811 | 1.873 |
| SISBLM [46] | 0.925 | 7.194 | 0.822 | 7.587 | 0.885 | 0.025 | 0.809 | 1.985 |
| ZHOU [47] | 0.951 | 5.747 | - | - | **0.919** | **0.018** | - | - |
| PROPOSED | **0.954** | **5.739** | **0.976** | **6.150** | 0.916 | 0.021 | **0.839** | **1.611** |

In order to demonstrate that the proposed method can also work on images with multiple distortions, it was also tested on images from four multiply distorted datasets LIVEMD [56], IVL MD [57], MDID2013 [46] and MDID [59] wherein original images were exposed to combination of JPEG compression, blurriness and white noise. The training and testing procedure for ML and Non-ML based methods was the same as described in section 4.1. The performance was compared to the state-of-art single-distortion (generic) methods BLIINDS2 [32], BRISQUE [29], GMLoG [33], NFERM [34], NIQE [41] and IL-NIQE [42] and multiple-distortion methods GWH-GLBP [44], Miao [45], and SISBLIM [46] using their source codes (on LIVEMD, IVL MD, MDID2013 and MDID) and results are listed in Table 8 and 9. For method proposed by Zhou [47], the scores were directly taken from original paper for LIVEMD and MDID2013. It is noteworthy to observe the relatively poor performance and inability of celebrated generic methods to assess multiple distortions in an image. From the correlation scores listed in Table 8 namely SROCC and KROCC, it can be analyzed that the proposed technique works well on all the databases for the combination of multiple distortions except for slightly inferior performance compared with Miao [45] and GWH-GLBP [44] on MDID. But at the same time, it outperforms Miao [45] when we compare the PLCC scores. The method was also compared using the PLCC and RMSE scores on all the four multiple distortion-based databases and are listed in Table 9. The proposed method performs better than GWH-GLBP [44] and Zhou [47] for mostly all the sub-sets. To study further, statistical analysis (t-test) was conducted for comparison of these methods on LIVEMD [56] dataset. NIQE [41] and IL-NIQE [42] were excluded as they are opinion unaware learning free methods. The numbers were used from 1 to 7 to represent methods NFERM [34], GWH-GLBP [44], BRISQUE [29], BLIINDS2 [32], GMLoG [33], Miao [45] and proposed method respectively. The symbols '1', '_', and '0' tell that the method in row is significantly better, similar or worse, compared to metric in corresponding column (with 95% confidence level). Table 10 represents these statistical analysis results wherein each symbol in every entry corresponds to results obtained on criteria's PLCC, SROCC and RMSE. It can be observed that the proposed method is statistically superior compared to all the benchmark techniques with the exception of Miao [45] for RMSE criteria.

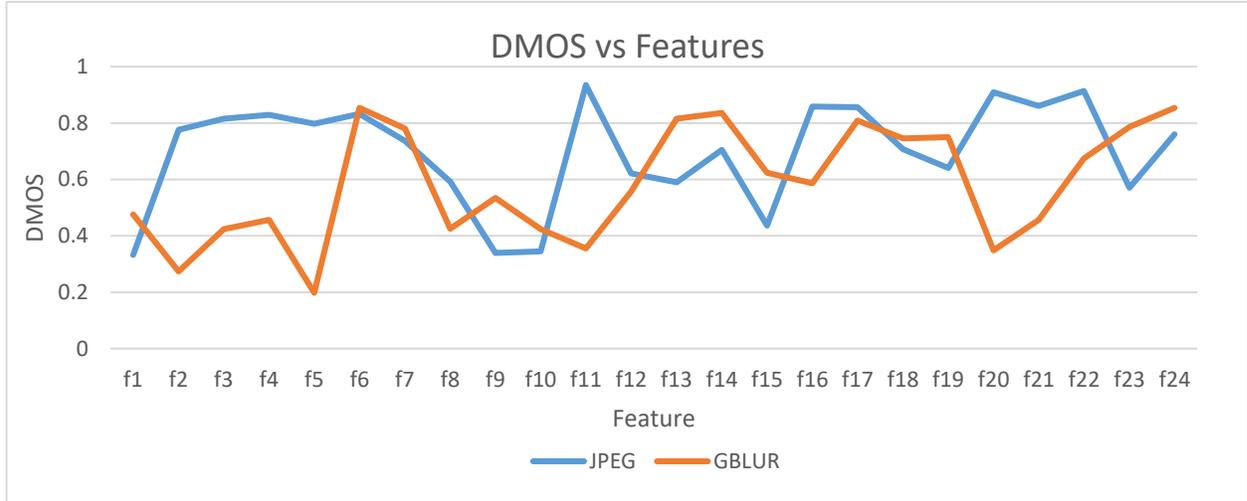

*Fig. 10 Correlation of features with human perceptions for SROCC scores on the two types of distortions*

**Table 10** Statistical Analysis between different methods for criteria's PLCC, SROCC and RMSE on LIVEMD database.

|   | 1 | 2 | 3 | 4 | 5 | 6 | 7 |
|---|---|---|---|---|---|---|---|
| 1 | ___ | ___ | 1__ | 111 | 111 | 000 | 000 |
| 2 | ___ | ___ | ___ | 111 | 111 | 000 | 000 |
| 3 | 0__ | ___ | ___ | 111 | 111 | 000 | 000 |
| 4 | 000 | 000 | 000 | ___ | 111 | 000 | 000 |
| 5 | 000 | 000 | 000 | 000 | ___ | 000 | 000 |
| 6 | 111 | 111 | 111 | 111 | 111 | ___ | __1 |
| 7 | 111 | 111 | 111 | 111 | 111 | __0 | ___ |

*4.5 Contribution of different features:*

To get a clear understanding of the relationship between the features and subjective scores we performed an experiment on LIVE [52] dataset consisting of all images present in GBLUR and JPEG subsets. Each feature was individually trained-tested using GPR learner for 80-20 % train-test ratio repeated over 1000 iterations to get all the possible combinations. The median scores are reported in Fig. 10, plotted separately for JPEG and GBLUR images. From this plot, it can be seen how zero-valued based features i.e. *f6*, *f11* & *f16* are highly relevant to human perception for GBLUR and JPEG images respectively. Similarly, as f21 and f22 are high-frequency based features hence show good results on JPEG compressed images. This work includes two parallel approaches one is only-DFT based and second is MSCN followed by DFT operation. To represent how this combination has better affinity towards assessment we mention the SROCC and PLCC scores for the three possible scenarios i.e. DFT, MSCN+DFT and Combination of DFT and MSCN+DFT in Table 11. It is very clear that the highest scores are obtained for the combination of DFT and MSCN+DFT uniformly across all three databases LIVE [52], CSIQ [5] and TID2013 [40].

*4.6 Choice of Regression model:*

To justify the use of GPR model as the mapping tool for extracted features, we make a comparison between the SROCC and PLCC scores obtained for Support Vector Machine (SVM) and GPR models respectively at two different validation ratios (80-20 and 50-50), as shown in Table 12. Clearly, GPR gives the best performance validating its employment in the model.

**Table 11** SROCC and PLCC values for three approaches. Best performance highlighted using bold face.

| Database | | LIVE | | TID2013 | | CSIQ | |
|---|---|---|---|---|---|---|---|
| Distortion Type | | JPEG | GBLUR | JPEG | GBLUR | JPEG | GBLUR |
| I. MSCN+DFT | SROCC | 0.964 | 0.958 | 0.904 | 0.911 | 0.903 | 0.918 |
| II. DFT | SROCC | 0.962 | 0.943 | 0.924 | 0.914 | 0.890 | 0.912 |
| Combination of I & II | SROCC | **0.979** | **0.972** | **0.958** | **0.951** | **0.948** | **0.944** |
| I. MSCN+DFT | PLCC | 0.971 | 0.965 | 0.918 | 0.922 | 0.926 | 0.906 |
| II. DFT | PLCC | 0.965 | 0.938 | 0.960 | 0.934 | 0.918 | 0.924 |
| Combination of I & II | PLCC | **0.986** | **0.971** | **0.966** | **0.953** | **0.949** | **0.941** |

**Table 12** Comparison of SROCC and PLCC values of IQA metrics based on the technique used. Best performance highlighted using bold face.

| Database | | LIVE | | TID2013 | | CSIQ | | Cross-Validation Ratio |
|---|---|---|---|---|---|---|---|---|
| Distortion Type | | JPEG | GBLUR | JPEG | GBLUR | JPEG | GBLUR | |
| GPR | SROCC | **0.979** | **0.972** | **0.958** | **0.951** | **0.948** | **0.944** | 80-20 |
| SVM | SROCC | 0.961 | 0.965 | 0.937 | 0.924 | 0.917 | 0.910 | 80-20 |
| GPR | SROCC | 0.975 | 0.962 | 0.953 | 0.933 | 0.926 | 0.928 | 50-50 |
| SVM | SROCC | 0.941 | 0.953 | 0.927 | 0.913 | 0.910 | 0.904 | 50-50 |
| GPR | PLCC | **0.986** | **0.971** | **0.966** | **0.953** | **0.949** | **0.941** | 80-20 |
| SVM | PLCC | 0.958 | 0.947 | 0.939 | 0.918 | 0.903 | 0.926 | 80-20 |
| GPR | PLCC | 0.974 | 0.959 | 0.953 | 0.942 | 0.933 | 0.927 | 50-50 |
| SVM | PLCC | 0.932 | 0.941 | 0.929 | 0.899 | 0.893 | 0.920 | 50-50 |

**Table 13** Run-time comparison for different methods (in seconds)

| Method | DIIVINE [31] | BRISQUE [29] | NFERM [34] | GMLoG [33] | FRIQUEE [37] | Miao [45] | Proposed |
|---|---|---|---|---|---|---|---|
| Time | 38.6 | 0.41 | 82.3 | 0.19 | 40.2 | 5.2 | 4.81 |

*4.7 Rum-time comparison:*

In order to measure and compare the run time-complexity, the proposed method and some of the existing methods such as DIVINE [31], BRISQUE [29], NFERM [34], GMLoG [33], FRIQUEE [37] and Miao [45] were executed on a computer equipped with Intel (R) Xeon (R) processor, CPU- 2.13 GHz., 20 GB usable RAM and 64-bit operating system and 1 TB HDD. The run time measured for each algorithm is shown in Table 13. These results are for a group of 100 image of different sizes chosen from LIVE database [52] of GBLUR and JPEG type distortions (50%-50%).

From Table 13, it can be observed that proposed algorithm gives a moderate performance for rum-time complexity. It gives slightly inferior performance (time-complexity) compared to BRISQUE [29] and GMLoG [33] but it is superior to DIIVINE [31], NFERM [34], FRIQUEE [37] and Miao [45]. However, it may be noted that although the run-time complexity for the proposed method is inferior to that of GMLoG [33] and BRISQUE [29] but the accuracy for the proposed method is higher compared to these two models. Hence, the model's combined performance and accuracy (SROCC, PLCC, KROCC and RMSE) along with time-complexity makes it the most suitable algorithm for assessing images distorted with either blocking artefacts, blurriness, noise individually or jointly.

## 5. CONCLUSIONS

In this work, a novel approach to estimate the quality of images corrupted with blocking artefacts, blurriness, or noise individually or jointly. The proposed model is inspired from human perceptions employing the transform domain DFT and MSCN coefficients to get the natural scene properties most suited to HVS. The proposed model effectively and accurately estimates the quality of images where a combination of blurriness, blocking artifacts and noise are observed. The proposed method is relatively faster and highly accurate. It is a fully blind method of image quality assessment and is capable of estimating the overall quality of images distorted jointly by blockiness, blurriness and noise, and useful to estimate quality of highly JPEG/MPEG/H.26x compressed images/videos received through bandlimited low-pass channels. This scenario is widely encountered while exchanging images through modern electronic gadgets. In future, we aimed to extend this method for VQA (video quality assessment) for real-time quality monitoring of HEVC and VVC coded videos.